\documentclass[letterpaper, 10 pt, journal]{ieeeconf} 

\usepackage{graphics} 
\usepackage{graphicx}   
\usepackage{amsmath} 
\usepackage{amssymb}  
\usepackage{mathtools, cuted}
\usepackage{interval}
\usepackage{nicefrac}
\usepackage[noadjust]{cite}
\usepackage{breqn}
\usepackage{color, colortbl}
\usepackage{derivative}
\usepackage{siunitx}
\usepackage{multirow}
\usepackage{hyperref}
\usepackage{cleveref}

\renewcommand{\vec}[1]{\mathbf{\boldsymbol{#1}}}

\newcommand{\etal}{\textit{et al}. }

\usepackage[dvipsnames]{xcolor}

\title{Harnessing Natural Oscillations for High-Speed, Efficient \\ Asymmetrical Locomotion in Quadrupedal Robots}

\author{Jing Cheng$^*$, Yasser G. Alqaham$^*$, and Zhenyu Gan
\thanks{All authors are with the Department of Mechanical and Aerospace Engineering, Syracuse University, Syracuse, NY 13244 \texttt{\{jcheng13, ygalqaha, zgan02\}@syr.edu}.}
\thanks{This work was supported by a startup fund from the Syracuse University.}
\thanks{$^*$The authors contribute equally to this paper.}
}

\begin{document}

\maketitle

\begin{abstract}
This study explores the dynamics of asymmetrical bounding gaits in quadrupedal robots, focusing on the integration of torso pitching and hip motion to enhance speed and stability. Traditional control strategies often enforce a fixed posture, minimizing natural body movements to simplify the control problem. However, this approach may overlook the inherent dynamical advantages found in natural locomotion. By considering the robot as two interconnected segments, we concentrate on stance leg motion while allowing passive torso oscillation, drawing inspiration from natural dynamics and underactuated robotics principles.
Our control scheme employs Linear Inverted Pendulum (LIP) and Spring-Loaded Inverted Pendulum (SLIP) models to govern front and rear leg movements independently. This approach has been validated through extensive simulations and hardware experiments, demonstrating successful high-speed locomotion with top speeds nearing 4 m/s and reduced ground reaction forces, indicating a more efficient gait. Furthermore, unlike conventional methods, our strategy leverages natural torso oscillations to aid leg circulation and stride length, aligning robot dynamics more closely with biological counterparts.
Our findings suggest that embracing the natural dynamics of quadrupedal movement, particularly in asymmetrical gaits like bounding, can lead to more stable, efficient, and high-speed robotic locomotion. This investigation lays the groundwork for future studies on versatile and dynamic quadrupedal gaits and their potential applications in scenarios demanding rapid and effective locomotion.
\end{abstract}

\section{Introduction}
Quadrupedal animals in nature typically resort to asymmetrical gaits such as bounding or galloping when moving at their peak speeds \cite{Hildebrand1977}. Unlike the symmetrical gaits like trotting, these asymmetrical patterns feature a rapid succession of footfalls, marked by significant body rotations and swift accelerations \cite{Gambaryan1975}. Inspired by these natural motions, numerous roboticists have sought to replicate such gaits in quadrupedal robots. Notably, Marc Raibert pioneered in this area, employing the concept of “virtual legs” and introducing bounding gait in robots equipped with telescoping legs as detailed in \cite{raibert1986legged}.
Further advancements were made by Poulakakis \etal that utilized straightforward control laws aided by template models to attain stable bounding, half-bounding, and rotary gallop in their Scout II robot \cite{Poulakakis2006}. Similarly, Park \etal explored impulse scaling, applying feedforward force profiles to facilitate automatic speed adaptation for bounding control in the MIT Cheetah2 \cite{Park2017}. 
These efforts underscore the meticulous crafting and tuning of desired motions and the reliance on heuristic feedback control laws to maintain gait stability.
\begin{figure}[tbp]
\centering
\includegraphics[width=1\columnwidth]{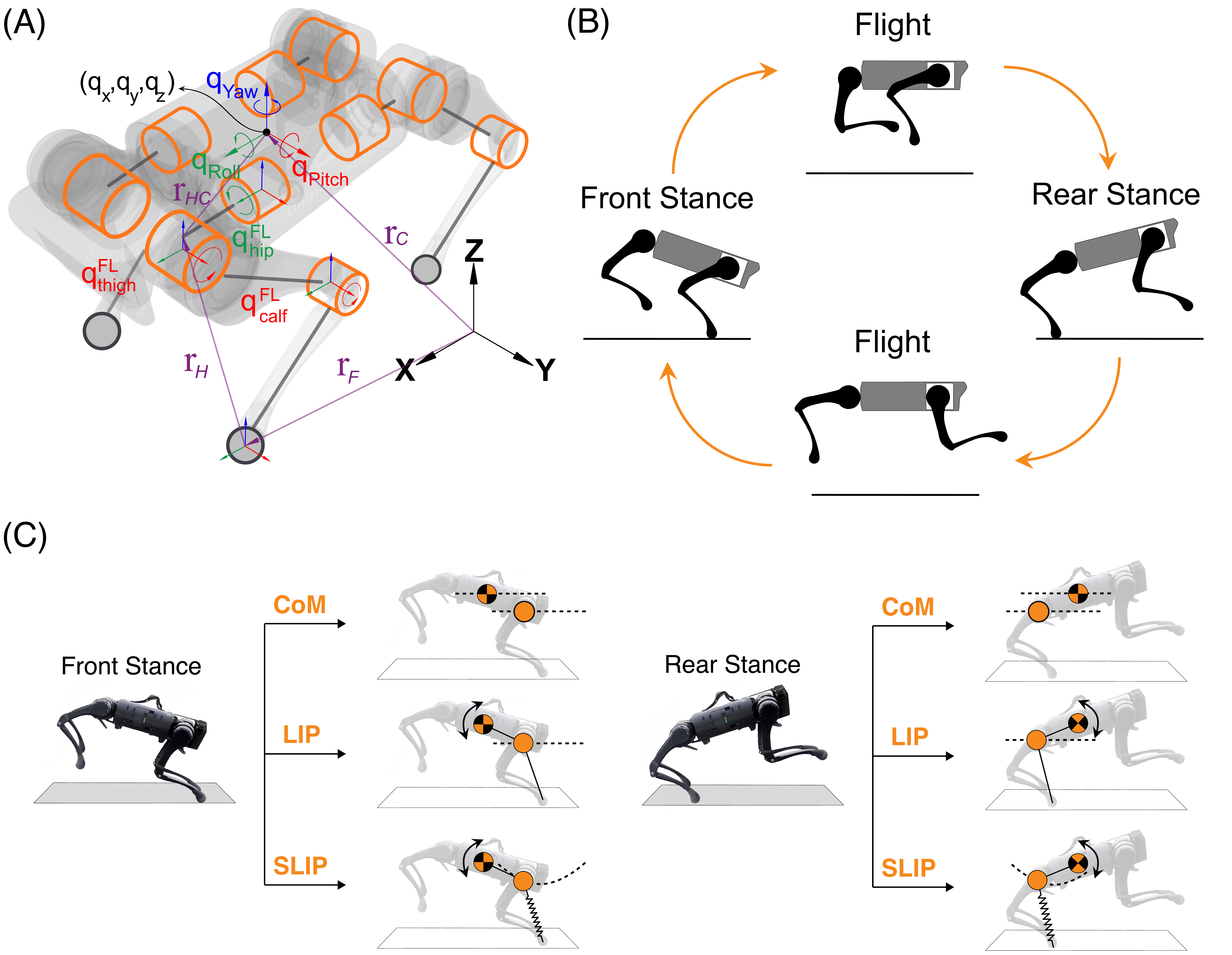}
\caption[Torque library]{(A) The configuration of quadrupedal robot A1 from Unitree Robotics. (B) The state machine of the bounding gait with two flight phases. (C) Three simplified models for the quadrupedal robot.}
\label{fig:RRF}
\vspace{-2mm}
\end{figure}

With advancements in portable computing and the development of efficient optimization algorithms, online trajectory planning for complex dynamical systems has become feasible. Techniques like Quadratic Programming (QP) and Model Predictive Control (MPC) have been developed to identify desired reference trajectories and devise optimal control strategies. These methods have proven effective in stabilizing robot movements across various gaits, even in unstructured environments or when facing significant disturbances such as kicks or unexpected pushes \cite{Minniti2022, Ding2019, kim2019highly}.
However, the effectiveness of these methods varies among different gaits when tasked with generating desired reference motions. For instance, to facilitate rapid replanning, many current approaches rely on simplified models, such as representing the system as a Single Rigid Body (SRB) \cite{kim2019highly} and assuming linear torso movement at a set speed. While this simplification yields satisfactory results in trotting gaits, where diagonal leg pairs strike the ground simultaneously, causing no net torque on the torso, it is less effective for asymmetrical bounding gaits. In bounding, where the gait cycle is divided into distinct stance phases for the front and rear leg pairs, torso rotation is an intrinsic part of the movement. Rather than suppressing the rotations of the torso motion, biological studies have shown that these rotational motions help with the circulations of the swing legs and enable longer stride lengths with improved energetic efficiencies \cite{cavagna1977mechanical, kim2014role}. Furthermore, our previous research on legged locomotion \cite{Gan2015Quadruped, GanDynamicSimilarity} has revealed that different gaits represent distinct oscillation modes within the same mechanical framework. It is most efficient to operate in harmony with these natural oscillation modes at their preferred frequencies. Conversely, compelling a mechanical system to move contrary to its inherent dynamics can lead to instability and increased energy expenditure.

Building on the principles of natural dynamics \cite{remy2011optimal, Gan2018allcommonbipedal} and underactuated robotics \cite{underactuated}, this study aims to propose and evaluate a control algorithm tailored for asymmetrical gaits in quadruped robots. Unlike conventional approaches that command the torso to maintain a straight, constant-speed trajectory with no rotation during stance phases, our method conceptualizes the robot as two interconnected segments. This approach focuses primarily on the motion control of the stance legs while facilitating natural oscillatory movements of the torso. In addition, it independently manages the motion of the leg pairs using established template models. The key contributions of this work are outlined as follows:
\begin{itemize}
  \item We have rigorously evaluated three types of control schemes: conventional method with no free torso rotation and direct hip motion control using two different template models. All approaches have been validated through high-fidelity simulations and experiments.
  \item The introduced methods have demonstrated the capability to sustain stable locomotion at high speeds, reaching up to nearly 4 m/s.
  \item The proposed methods have been effective in diminishing the magnitude of required reaction forces, indicating a more efficient gait at elevated speeds.
\end{itemize}

The remainder of the paper is structured as follows:
Section \ref{sec:methods} elaborates on the detailed design and structure of the proposed control scheme. Section \ref{sec:results} assesses the performance of our approach, specifically applied to a bounding gait with dual flight phases, highlighting the utilization of template models for each leg and benchmarking against the conventional CoM driving method. Finally, Section \ref{sec:conclusions} provides conclusions drawn from the study.

\section{Methods}
\label{sec:methods}
This section provides a detailed description of the A1 quadruped robot's structural model and delineates the proposed architecture for bounding gait control.

\subsection{Robot Model}
\label{sec:Robot Model}
The A1 quadrupedal robot from Unitree possesses a total of 18 degrees of freedom (DOF), each leg being equipped with three revolute joints, actuated by three electric motors. The main body's position is articulated in Cartesian coordinates $(q_x, q_y, q_z)$ with respect to the inertial frame. The orientation of the main body, relative to the same frame, is parameterized by Euler angles $(q_{\text{yaw}}, q_{\text{pitch}}, q_{\text{roll}})$. The angles for the $i$th leg's joints $ \vec{q}^i_{\text{leg}}  \mathrel{\mathop:}= [q_{\text{hip}}^i, q_{\text{thigh}}^i, q_{\text{calf}}^i ]^\intercal$ represent, respectively, the hip's angle relative to the torso, the thigh's angle in relation to the hip, and the calf's angle with respect to the thigh, adhering to the right-hand rule convention. The index $i$ spans the set $\{\text{FR}, \text{FL}, \text{RR}, \text{RL}\}$, denoting front-right, front-left, rear-right, and rear-left legs, respectively.

The floating-base model's configuration space $\mathcal{Q}$ for A1 is encapsulated by the generalized coordinates vector $\vec{q}$, integrating all configuration variables into a unified representation. The motion equations for the A1 robot, formulated through the Euler-Lagrange equation, are presented as follows:
\begin{equation}
\label{eq:EOM}
   \operatorname{\vec{M}}(\vec{q})\ddot{\vec{q}} + \operatorname{\vec{C}}(\vec{q},\dot{\vec{q}})\dot{\vec{q}} + \operatorname{\vec{G}}(\vec{q}) = \vec{S}\vec{\tau}  + \sum_{i }\vec{J}_{i}^\intercal(\vec{q})\vec{f}_i.
\end{equation}
Here, $\operatorname{\vec{M}}(\vec{q}) \in \mathbb{R}^{18 \times 18}$ denotes the mass-inertia matrix, $\operatorname{\vec{C}}(\vec{q},\dot{\vec{q}}) \in \mathbb{R}^{18}$ the Coriolis and centrifugal force matrix, and $\operatorname{\vec{G}}(\vec{q}) \in \mathbb{R}^{18}$ the gravitational force vector. The vector $\vec{\tau} \in \mathbb{R}^{12}$ represents the joint torques, with $\vec{S} = \left[\vec{0}_{6 \times 12}; \vec{I}_{12 \times 12}\right]$ acting as the selection matrix, correlating motor torques to their corresponding joints. Furthermore, assume the $i$-th foot is in stance and its location in the inertial frame is $\vec{g}_l(\vec{q})$, the transpose of the contact Jacobian matrix $\vec{J}_i(\vec{q})$ effectively maps the corresponding ground reaction forces (GRFs) $\vec{f}_i \in \mathbb{R}^{3}$ onto the generalized coordinates.

\subsection{Bounding Control Design}
\label{sec:Bounding Control Design}
\begin{figure*}[t!]
\centering
\includegraphics[width=2\columnwidth]{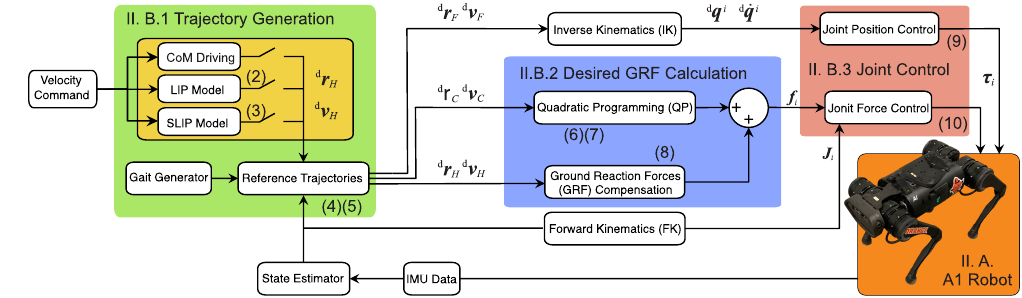}
\caption[COT]{This figure depicts the proposed control framework, with sections on trajectory generation, desired GRF calculation, and joint control highlighted in green, blue, and red, respectively. Detailed calculations for each module are annotated with corresponding equation numbers from the text.}
\label{fig:Control}
\vspace{-2mm}
\end{figure*}
The bounding gait exhibits several variations depending on the number of aerial suspensions and the corresponding sequence of footfalls. Despite these complexities, they are characterized by distinct stance phases, supported solely by either the front or rear legs. Such configuration invariably induces torso rotations in the sagittal plane, denoted as $q_{\text{pitch}}$. Rather than commanding the torso to counteract the pitching motion, our controller design opts to focus on managing the motion through the hip joints of the stance leg pair, allowing the torso to rotate freely. The entire control structure is delineated in Fig.~\ref{fig:Control}, with detailed explanations provided in the subsequent subsections.

\subsubsection{Trajectory Generation}
\label{sec:TG}
The generation and online replanning of reference trajectories for a hybrid system with high degrees of freedom, as described in equation~\eqref{eq:EOM}, represent significant challenges. To achieve agile and stable behavior, we adopt the approach proposed in \cite{di2018dynamic}, approximating the robot as a single rigid body influenced by GRFs at the foot contacts. This method allows for the generation of desired torso accelerations at a high refresh rate. In this study, three methods have been utilized to generate desired torso motions, as delineated in Fig.~\ref{fig:Control}:

\paragraph{CoM (Center of Mass) driving}
\label{par:Straight_Line}
Adhering to conventional practices, as observed in the Mini Cheetah \cite{kim2019highly} and A1 robot \cite{Sombolestan2021}, we assume there is no body rotation and during the stance phase that the hip joint progresses forward in a straight line, maintaining a constant height. 

\paragraph{Linear Inverted Pendulum (LIP) Model}
The LIP model, widely recognized for its utility in legged locomotion, posits that there is no vertical oscillation of the system during the stance phase. Instead, the horizontal motion exhibits deceleration and acceleration akin to that of a pendulum, governed by:
\begin{equation}
\label{eq:LIP}
\ddot{r}_H^x = \frac{k g}{r_H^z} (r_H^x - r_F^x),
\end{equation}
where the hip joint is presumed to move similarly to a LIP with a constant height $r_H^z$, $k$ is a positive constant, $g$ represents gravity, and $r_H^x$ and $r_F^x$ denote the horizontal positions of the hip and stance foot, respectively.

\paragraph{Spring-Loaded Inverted Pendulum (SLIP) Model}
Contrary to the fixed-height assumption of the LIP model, the SLIP model better encapsulates the dynamic motions at higher speeds, accounting for leg behaviors such as shortening, compression, thrust, and extension, akin to a single linear spring. Owing to the absence of explicit analytical solutions, we approximate the hip joint's motion during the stance phase using sinusoidal dynamics:
\begin{equation}
\label{eq:hip_motion}
\begin{aligned}
r_H^z = & -\left(0.01 + 0.0025 \, v_C^x \right) \sin\left(\frac{\pi}{{T}{\gamma}} t_{s}\right) \\
&+ \left( 0.3 - 0.01 v_C^x \right),
\end{aligned}
\end{equation}
where $v_C^x$ denotes the horizontal speed of the CoM, $T$ the total stride duration, $\gamma$ the duty factor, and $t_s$ the elapsed time since the start of the current stance phase. The parameters are determined through curve fitting, accommodating the velocity dependencies indicated by the SLIP model, such as increasing angle of attack and spring compression correlating with higher forward speeds.

\vspace*{0.1in}
Upon determining the hip motion via one of the specified models, it is crucial to define the desired motion trajectory for the robot's torso CoM. In the case of the CoM driving model, we align with the methodology outlined in \cite{Bledt2018} by assuming the torso maintains a non-rotational stance during movement, thereby ensuring a consistent motion trajectory across all points on the torso, including the CoM. In contrast, for the LIP and SLIP models, which allow for torso rotation, the desired motion of the torso can be inferred from the established kinematic equations:
%
\begin{align}
{}^{\text {d}}\vec{r}_{C} &= {}^{\text {d}}\vec{r}_{H} + \vec{r}_{H C}, \\
{}^{\text {d}}\vec{v}_{C} &= {}^{\text {d}}\vec{v}_{H} + \vec{\omega} \times \vec{r}_{H C}, 
\end{align}
where the superscript ${}^{\text {d}}( \cdot )$ denotes desired reference values, and $\vec{r}_{HC}$ represents the vector of relative displacement between the hip and the CoM.

\vspace*{0.1in}
\subsubsection{Desired GRF calculation}
\paragraph{Quadratic Programming}
Upon determining the desired motion for the CoM, the desired GRFs are computed by solving a Quadratic Program (QP). This QP formulation simplifies the system dynamics by treating the entire structure as a single rigid entity and determines the optimal force vectors, subject to constraints that ensure these forces fall within the limits defined by the friction cone.
The optimization employs the following cost function to minimize both the magnitudes of the forces and the deviations from their values in the preceding iteration:
\begin{align}
\label{eq:cost_function}
J &= \frac{1}{2} \boldsymbol{f}^{\text{T}}\left(\boldsymbol{A}^{\text{T}} \boldsymbol{S} \boldsymbol{A} + \alpha \boldsymbol{W} + \beta \boldsymbol{U}\right) \boldsymbol{f}  \\ \notag
&+ \left(-\boldsymbol{b}_d^{\text{T}} \boldsymbol{S} \boldsymbol{A} - \boldsymbol{f}_{\text{pre}}^{\text{T}} \beta \boldsymbol{U}\right) \boldsymbol{f}, \\ \notag
   \text{where} \quad \footnotesize \boldsymbol{A} &= 
    \begin{bmatrix} 
        \boldsymbol{1} & \boldsymbol{1} & \boldsymbol{1} & \boldsymbol{1} \\ {\left[\vec{r}_{FC}^{\text{FR}}\right]_{\times}} & {\left[\boldsymbol{\vec{r}}_{FC}^{\text{FL}}\right]_{\times}} & {\left[\boldsymbol{\vec{r}}_{FC}^{\text{RR}}\right]_{\times}} & {\left[\boldsymbol{\vec{r}}_{FC}^{\text{RL}}\right]_{\times}}
    \end{bmatrix} \\ \notag
    \text{and} \quad \footnotesize  \boldsymbol{b}_d &= 
    \begin{bmatrix} 
        m(\dot{\boldsymbol{v}}_C - \boldsymbol{g}) \\ 
        \vec{R} \boldsymbol{I} \vec{R}^{\text{T}} \dot{\boldsymbol{\omega}} 
    \end{bmatrix}. \notag
\end{align}
%
$\vec{R}$ is the rotation matrix transitioning from the body-fixed coordinate system to the world frame. $m$ denotes the mass of the simplified model, and $\boldsymbol{r}_{FC}^i$ represents the displacement from foot to CoM within the world coordinate frame. $\boldsymbol{f} \in \mathbb{R}^{12}$ encapsulates the GRFs $\boldsymbol{f}_{i} \in \mathbb{R}^{3}$ for each leg, while $\boldsymbol{I}$ signifies the simplified torso's inertia tensor. $\boldsymbol{S}$, $\boldsymbol{W}$ and $\boldsymbol{U}$ are adjustable weight matrices. The constants $\alpha$ and $\beta$ are scalar coefficients, respectively, and $\boldsymbol{f}_{\text{pre}}$ references the GRFs computed in the preceding iteration.
The GRF constraints are implemented as follows:
\begin{equation}
\vec{M}_{\mu} \boldsymbol{f} \geqslant \mathbf{0},
\end{equation}
where $\vec{M}_{\mu}$ represents the matrix encoding the static friction coefficients, ensuring that the feasible region for $\boldsymbol{f}$ forms a quadrangular pyramid. This QP approach has been successfully implemented by Unitree Robotics, enabling the A1 robot to perform trotting, walking, and crawling gaits \cite{Unitree_guide}.

\paragraph{GRF Compensation}
The simplifications employed in the QP, particularly at high velocities, necessitate additional vertical GRFs to enhance the hip tracking performance. This is achieved through the following feedback control mechanism for each stance leg:
%
\begin{equation}
\label{eq:GRFCompen}
f_{i}^z = k_{\text{p}}({}^{\text {d}}r^{z}_{H} - r^{z}_{H}) + k_{\text{d}}({}^{\text {d}}v^{z}_{H} - v^{z}_{H}),
\end{equation}
%
where $k_{\text{p}}$ and $k_{\text{d}}$ represent the proportional and derivative feedback coefficients, respectively. This compensatory adjustment is integrated into the desired GRFs derived from the QP before being transmitted to the joint-level controller, as illustrated in Fig.~\ref{fig:Control}.

\vspace*{0.1in}
\subsubsection{Joint Control}
\paragraph{Joint Position Control}
During the stance phase, with the foot anchored to the ground, the desired hip positions and velocities, ${{}^{\text{d}}\vec{r}}_{H}$ and ${{}^{\text{d}}\vec{v}}_{H}$, are established and subsequently converted into desired joint positions via inverse kinematics (IK). In the swing phase, our primary aim is accurate foot placement at predetermined footholds. Here, the desired joint angles and velocities are computed from the specified foot positions and velocities, ${{}^{\text{d}}\vec{r}}_{F}$ and ${{}^{\text{d}}\vec{v}}_{F}$, based on Marc Raibert's law of foot placement \cite{raibert1986legged}.

\paragraph{Joint Force Control}
For the swinging leg, once the target foot position is identified, Proportional-Derivative (PD) control is employed to exert a virtual corrective force ensuring adherence to the intended trajectory:
%
\begin{equation}
\label{eq:flightForce}
\vec{f}_i^{\text{sw}} = \vec{K}_{\text{p}}\left({}^{\text{d}}\vec{r}^i_{F} - \vec{r}^i_{F}\right) + \vec{K}_{\text{d}}\left({}^{\text{d}}\vec{v}^i_{F} - \vec{v}^i_{F}\right),
\end{equation}
%
where $\vec{K}_{\text{p}}$ and $\vec{K}_{\text{d}}$ are positive definite matrices denoting proportional and derivative gains, respectively.

For stance legs, subsequent to the QP algorithm determining the desired GRFs, $\boldsymbol{f}^{\text{st}}_{i}$, and under the assumption of massless leg segments, the requisite joint forces across hip, thigh, and calf are computed:
%
\begin{equation}
\label{eq:force_mapping}
\vec{\tau}_{i} = -\vec{J}_{i}^{\text{T}} \boldsymbol{f}^{\text{sw/st}}_{i},
\end{equation}
%
where $\vec{J}_{i}^{\text{T}}$ is the Jacobian transpose, facilitating the conversion from leg joint movements to foot position, defined as $\vec{J}_{i}^{\text{T}} \equiv \pdv{\vec{g}^i}{\vec{q}^i_{\text{leg}}} \in \mathbb{R}^{3 \times 3}$, with $\vec{g}^i$ indicating the foot's location in the inertial frame. The torque vector for the $i$-th leg is denoted by $\vec{\tau}_i$.

\section{Simulation and Hardware Results}
\label{sec:results}
In this section, we compare the simulation outcomes for bounding gait reference trajectory generation at different speeds using three methodologies: the CoM driving method, controlling with the LIP model, and controlling with the SLIP model. Simulations were executed in ROS \cite{ROS} and Gazebo \cite{koenig2004design}. Hardware validations on the A1 robot, utilizing LIP and SLIP models, confirmed the control algorithms' effectiveness with comparable results.
\subsection{Simulation Results}
\begin{figure}[t!]
\centering
\includegraphics[width=1\columnwidth]{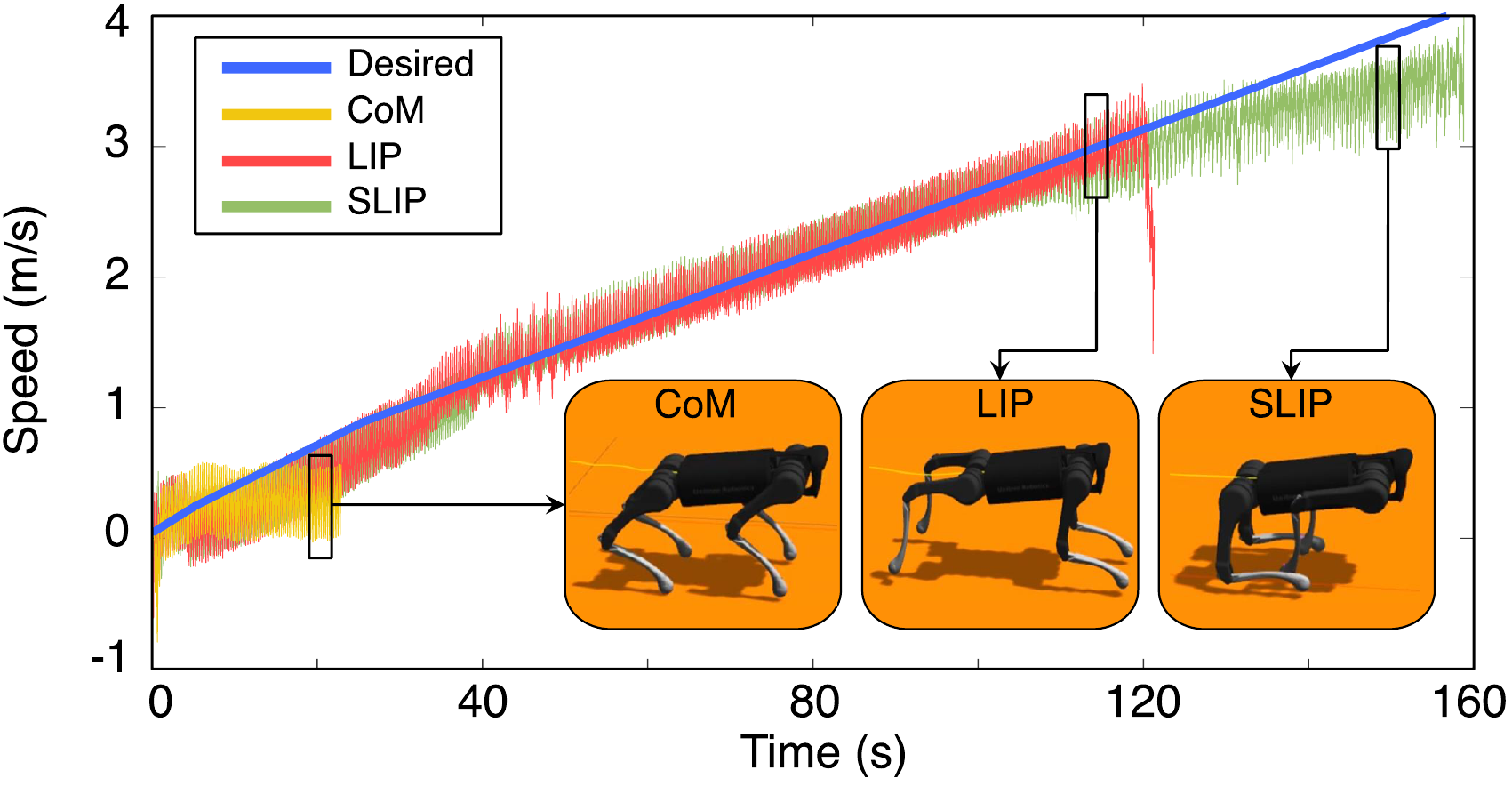}
\caption[COT]{This figure illustrates the maximum speed the robot can reach by each control strategy. The three subplots respectively show the posture of the robot when it reaches the speed in the corresponding black box.}
\label{fig:SimulationSpeed}
\vspace{-2mm}
\end{figure}
%
\begin{figure*}[tbp]
\centering
\includegraphics[width=2\columnwidth]{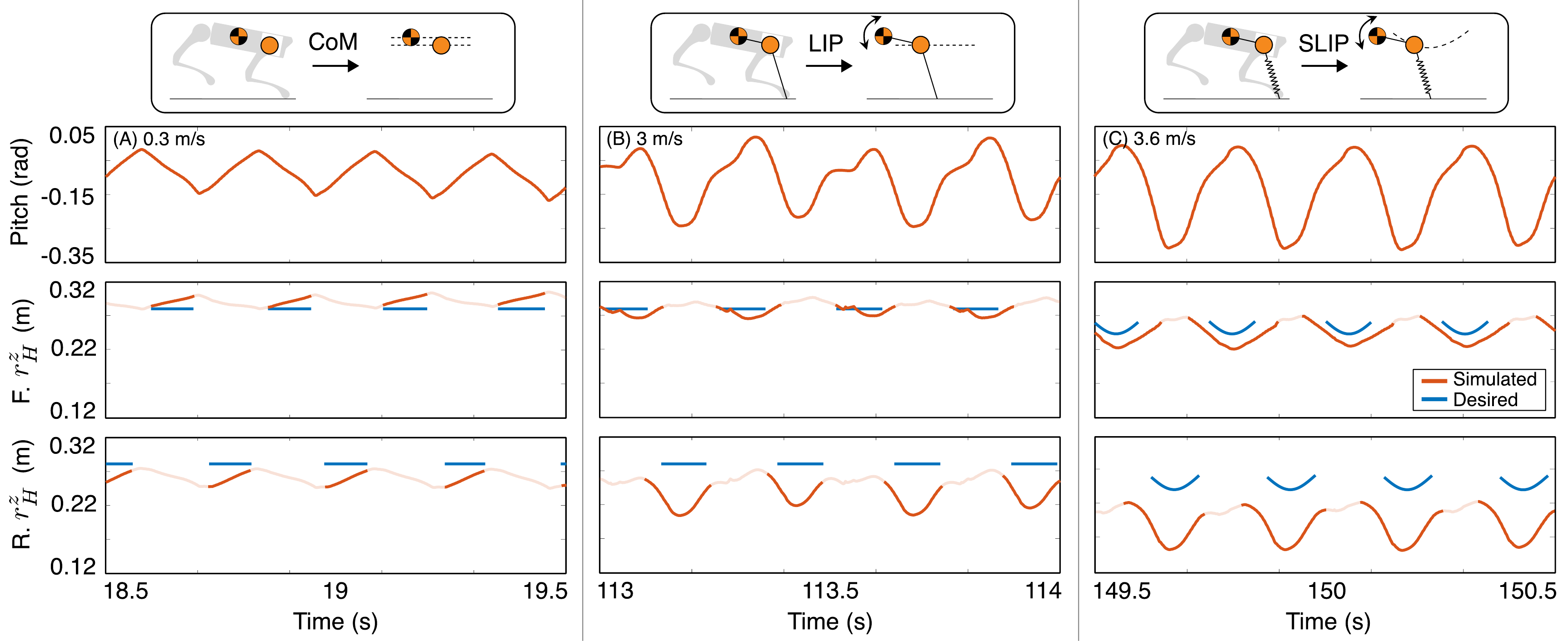}
\caption[Torque library]{The trajectory tracking performances of the A1 robot in the simulation are compared among three types of control schemes in this figure. As shown in (A), (B), and (C), the torso's pitching motion is unavoidable, and guiding its motion smoothly is essential. Different simplified models result in distinctly characteristic hip motions in the robot: The results with CoM driving demonstrate a quick change in pitching motion, whereas the other two solutions exhibit a more gradual adjustment. In hip motion, the first two solutions tend to maintain the desired height, while the control scheme with the SLIP model shows the actual trajectory moving along the envelope of the desired trajectory. The control outcomes align with our simplified model expectations. In all figures, the reference trajectories are shown in blue lines, and the actual trajectories are in red lines.}
\label{fig:SimulationResult}
\vspace{-2mm}
\end{figure*}
%
\begin{figure*}[tbp]
\centering
\includegraphics[width=2\columnwidth]{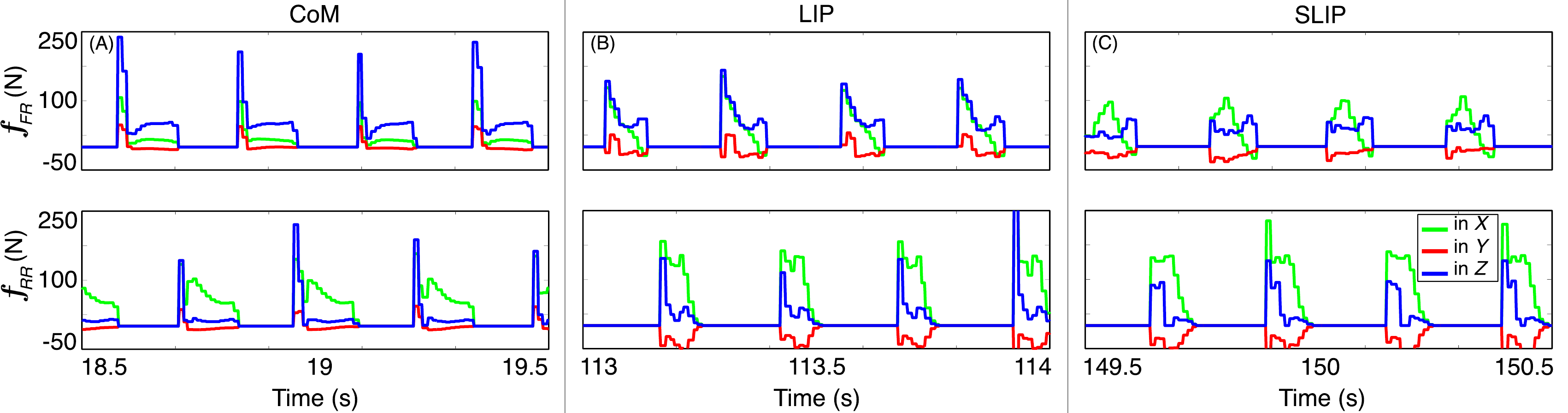}
\caption[Torque library]{This figure displays the GRF profiles under three control schemes at their maximum speeds, as indicated in Fig.~\ref{fig:SimulationSpeed}.}
\label{fig:SimulationGRF}
\vspace{-2mm}
\end{figure*}
%
Three distinct methodologies utilizing the full-body model of the A1 robot were assessed in the simulation. Post-initialization, each model directed the robot to accelerate from 0 m/s to 4 m/s, as depicted in Fig.~\ref{fig:SimulationSpeed}. Under the CoM driving approach, the robot maintained stable bounding sequences but failed to exceed speeds of 0.3 m/s, ceasing to adhere to velocity commands beyond this point. Conversely, the LIP model significantly outperformed the CoM method, accurately following the speed command up to 3.0 m/s until a loss of contact with the ground led to a fall. The SLIP model proved superior within these tests, enabling the robot to approach speeds of nearly 4 m/s. However, at such velocities, mechanical interferences arose as the legs converged beneath the torso during aerial phases, as illustrated in Fig.~\ref{fig:SimulationSpeed}(C). Inspired by nature's Cheetah galloping gaits, modifications were made by reducing the y-axis distance between the front feet and increasing it for the rear, enhancing gait stability. Despite these adjustments, precise speed command adherence was compromised at higher velocities. A closer investigation at the top speeds of all three cases was shown in Fig.~\ref{fig:SimulationResult} and Fig.~\ref{fig:SimulationGRF}.

\subsubsection{CoM Driving}
In this case, at a speed of 0.3 m/s, despite maintaining a constant desired pitching angle throughout the stride, the torso exhibited a zig-zag rotation due to asymmetric torque application during the stance phases, with an average deviation of 0.1 rad. Analysis of the front and rear hip joint tracking revealed that, while the QP effectively generated the desired GRFs for torso stability, there was a lack of coordination between the front and rear leg pairs regarding speed adherence: the robot accelerated during front stance and decelerated during rear stance, leading to an eventual equilibrium at a reduced speed, diverging from the intended velocity command. When looking at the corresponding GRFs in Fig~\ref{fig:SimulationResult}, large impulses were observed in the vertical directions (z-axis). This is due to the largest tracking errors observed in the pitching angle at the beginning of both front and rear touch-down events, and the QP immediately generated a large torque with the priority to correct pitching motion.

\subsubsection{LIP}
In the scenario employing the LIP model approach, the torso exhibited smoother rotations with a larger average magnitude of approximately 0.15 rad. Despite maintaining a constant vertical height for the hip joint, significant vertical oscillations were noted at peak speeds (3.0 m/s): the hip joints descended, compressing the legs in the first half of the stance phase before ascending, thereby reinstating the hips to their initial heights.
When contrasted with the CoM driving scenario, a noticeable reduction in the peak values of vertical GRFs at the start of each stance phase was evident within the LIP model approach, alongside smoother transitions observed in the torque profiles.

\subsubsection{SLIP}
Employing the SLIP model induced pronounced pitching oscillations, reaching up to 0.3 rad at peak robot speeds, as illustrated in Fig.~\ref{fig:SimulationResult}. These increased oscillations, particularly beneficial at high speeds, facilitated more effective leg circulation during the swing phase and consequently increased the stride length. The reference trajectories, designed to include vertical oscillations at the hip joints, led to a marked parabolic movement pattern, suggesting a greater extent of leg compression.
A noteworthy observation, akin to the LIP model scenario, is the disparity in hip tracking accuracy between the front and rear leg pairs, despite identical reference trajectory commands for both. The tracking performance of the rear leg pair's hip was consistently inferior to that of the front, with the rear hip's stance phase height registering lower. This discrepancy contributed to a skewed pitching motion in the torso, averaging an angle of -0.16 rad.
Moreover, the sophisticated design of these reference trajectories yielded more effective GRF profiles as velocities neared 3.6 m/s, as depicted in Fig.~\ref{fig:SimulationGRF}. Remarkably, compared to earlier models, the vertical GRFs were diminished, showcasing nearly smoother transitions, especially noticeable in the vertical forces applied by the front leg pair.

\subsection{Hardware Validation}
\label{sec:hardware_results}
%
\begin{figure}[tbp]
\centering
\includegraphics[width=1\columnwidth]{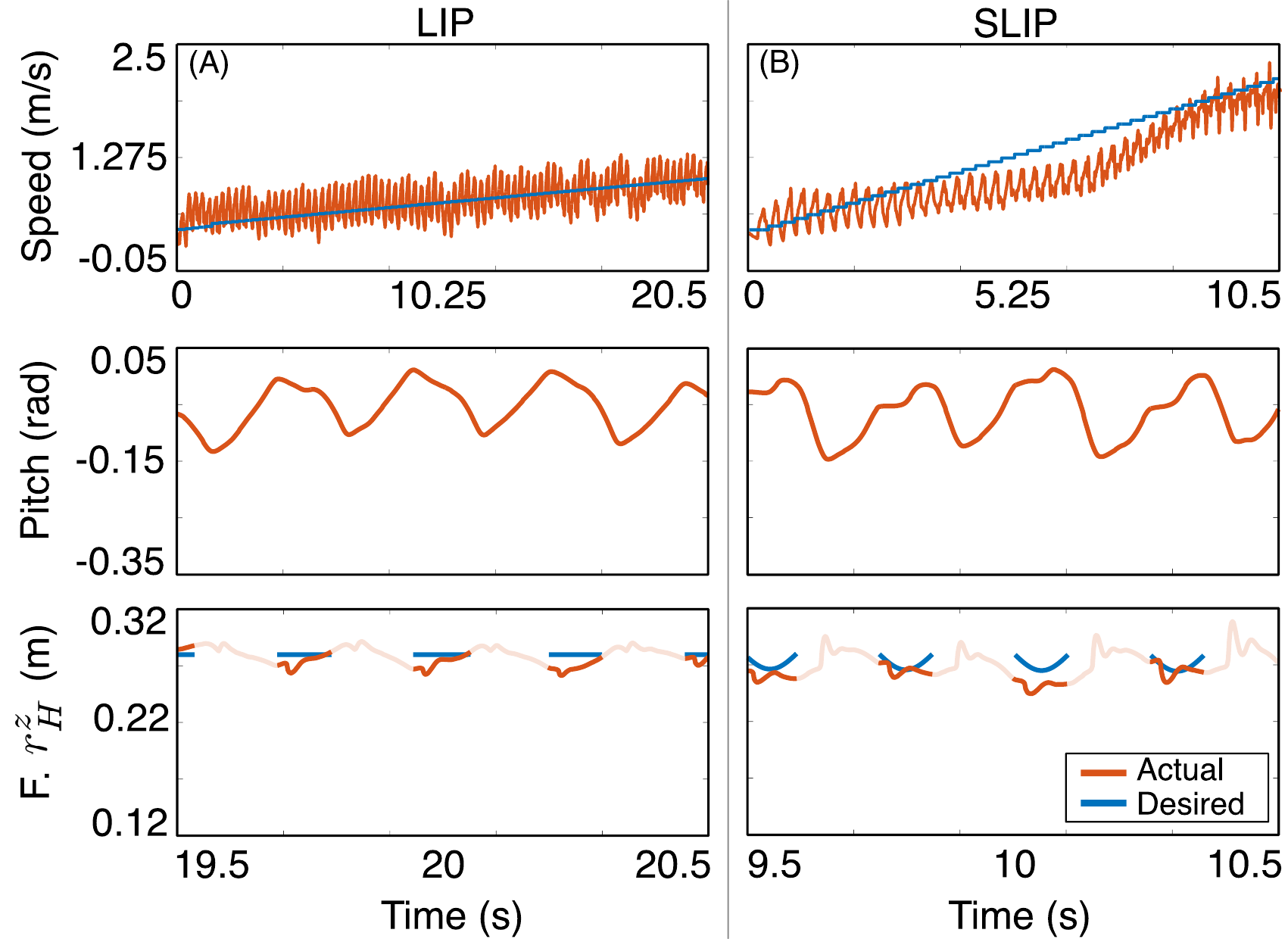}
\caption[Torque library]{This figure compares the experimental outcomes of control using LIP and SLIP models, revealing trends consistent with those observed in the simulation.}
\label{fig:experiments}
\vspace{-2mm}
\end{figure}
%
To evaluate the practical efficacy of the proposed control strategies, we replicated the tests on hardware, progressively increasing the robot's forward speeds. The effectiveness of both the LIP and SLIP methods on actual hardware is demonstrated in the accompanying multimedia file \footnote{\href{https://github.com/DLARlab/HarnessingNaturalOscillationsRobot}{https://github.com/DLARlab/HarnessingNaturalOscillationsRobot}}.
Constraints due to the hardware being tethered to external cables limited our testing space, preventing the robot from achieving its top speed before needing to deactivate the controller. Future work includes conducting these tests on a treadmill to bypass spatial limitations.
Overall, when compared to simulated outcomes, experimental data reflected significant torso rotations as depicted in Fig.~\ref{fig:experiments}. Although top speeds observed in simulations were not achieved in the hardware tests—reaching maximums of approximately 0.6 m/s for LIP and 2.0 m/s for SLIP. And lesser tracking discrepancies in hip movements were noted. Despite these variances, the overarching trends remained consistent with simulation predictions, with the robots moving dynamically and maintaining stability.

%
%
%
%

\section{Conclusions}
\label{sec:conclusions}
This paper introduces a control scheme specifically designed for high-speed, asymmetrical bounding gaits in quadrupedal robots. Unlike traditional bounding controllers that maintain a zero pitching angle in the torso (such as the CoM driving method), our strategy acknowledges the interplay between the front and rear leg pairs, allowing the robot's torso to rotate passively without direct intervention.
We evaluated two simplified models, LIP and SLIP, for independent control of the front and rear hip motions during the stance phases. Both models facilitated high-speed, stable locomotion. Instead of inhibiting the torso's pitching motion, we observed that allowing passive pitching enhanced the swing legs' circulation and notably decreased the necessary GRFs, even at elevated speeds.
Furthermore, the SLIP model demonstrated superior performance due to its more pronounced vertical oscillations in the stance phases, which additionally diminished the required GRFs, particularly in the initial half of the stance phase.
Future research will explore extending these control strategies to other asymmetrical gaits, including half-bounding and galloping \cite{ding2023breaking}, and will assess the robot's performance on unstructured terrains.

\bibliographystyle{IEEEtran}
\bibliography{References}

\end{document}